\documentclass[twoside,leqno,twocolumn]{article}

\usepackage[letterpaper]{geometry}

\usepackage{ltexpprt}
\usepackage{hyperref}
\usepackage{graphicx}
\usepackage{caption}
\usepackage{subcaption}
\usepackage{float}
\usepackage{amsmath}
\usepackage{amssymb}
\usepackage{comment}
\usepackage{color}
\usepackage{soul}

\usepackage{ulem}
\usepackage{xspace}
\usepackage[usenames,dvipsnames]{xcolor} 
\usepackage[affil-it]{authblk}

\begin{document}

\title{\Large Graph Sparsifications using Neural Network Assisted Monte Carlo Tree Search}
\author{Alvin Chiu$^1$, Mithun Ghosh$^2$, Reyan Ahmed$^2$, Kwang-Sung Jun$^2$, \\Stephen Kobourov$^2$, Michael T. Goodrich$^1$}
\affil{$^1$University of California, Irvine, CA, USA\\
$^2$University of Arizona, Tucson, AZ, USA}

\date{}

\maketitle







\begin{abstract} \small\baselineskip=9pt 
Graph neural networks have been successful for machine learning, as well as for combinatorial and graph problems such as the Subgraph Isomorphism Problem and the Traveling Salesman Problem.
We describe an approach for computing graph sparsifiers by combining a graph neural network and Monte Carlo Tree Search. We first train a graph neural network that takes as input a partial solution and  proposes a new node to be added as output. 
This neural network is then used in a Monte Carlo search to compute a sparsifier. The proposed method consistently outperforms several standard approximation algorithms on different types of graphs and often finds the optimal solution.
\end{abstract}

\section{Introduction}
Graph representation of data is a powerful approach to hold the relationships between different objects. Graphs arise in many real-world applications that deal with relational information.
Classical machine learning models, such as neural networks and recurrent neural networks, do not naturally handle graphs. Graph neural networks (GNN) were introduced 
to better capture graph structures~\cite{scarselli2008graph}. A GNN is a recursive neural network where nodes are treated as state vectors and the relationships between the nodes are quantified by the edges. 

Many real-world problems are modeled by combinatorial and graph problems that are known to be NP-hard.
GNNs offer an alternative to traditional heuristics and approximation algorithms; indeed the initial GNN model~\cite{scarselli2008graph} was used to approximate solutions to two classical graph problems: subgraph isomorphism and clique detection.


Recent GNN work~\cite{li2018combinatorial,xing2020graph} suggests that combining neural networks and tree search leads to better results than neural networks alone. Li et al.~\cite{li2018combinatorial} combine a convolutional neural network with tree search to  compute independent sets and other NP-hard problems that are efficiently reducible to the independent set problem. AlphaGo~\cite{xing2020graph} 
combines deep convolutional neural networks and Monte Carlo Tree Search (MCTS) 
to assess Go board positions and reduce the search space. 
Xing et al.~\cite{xing2020graph} used the same framework to tackle the traveling salesman problem (TSP).

Since Xing et al.~\cite{xing2020graph} showed that the AlphaGo framework is effective for TSP, a natural question is whether this framework can be applied to other combinatorial problems such as different graph sparsification problems~\cite{ahmed2023multi}. The Steiner tree and graph spanners are some examples of graph sparsification. Although these graph sparsification problems are NP-hard similar to TSP, there are several major differences among the natures of these problems. First, the sparsification problems contain a subset of the nodes called {\it terminals} that must be spanned, whereas in TSP all nodes are equivalent. Second, the output of the sparsification problem is a subgraph, whereas the output of TSP is a path (or a cycle). When iteratively computing a TSP solution, the next node to be added can only be connected to the previous one, which is much easier than having to choose from a set of nodes when growing a sparsification.
Third, TSP and Go are similar in terms of the length of the instance: both the length of the game and the number of nodes in the TSP solution are fixed, and taking an action in Go is equivalent to adding a node to the tour, while the number of nodes in the sparsification problem varies depending on the graph instance. 
Finally, Xing et al.~\cite{xing2020graph} only considered geometric graphs, which is a restricted class of graphs.

\subsection{Background:}
A sparsification of a graph $G$ is a subgraph that preserves some properties of $G$~\cite{ahmed2023multi}. Examples of sparsifications include spanning trees, Steiner trees, spanners, and distance preservers. Many sparsification problems are defined with respect to a given subset of vertices $T \subseteq V$ which we call terminals: e.g., a Steiner tree over $(G, T)$ requires a tree in $G$ which spans $T$. 

The Steiner tree problem is a classical NP-hard problem~\cite{ahmed2019multi}. In this problem, we are given an edge-weighted graph $G=(V,E)$, and a set of terminals $T\subseteq V$. And we want to compute a minimum weighted subtree that spans all terminals. For $|T|=2$ this is equivalent to the shortest path problem, for $|T|=|V|$ this is equivalent to the  minimum spanning tree problem, while the problem is NP-hard for $2<|T|<|V|$~\cite{cormen2009introduction}. 
Due to applications in many domains, there is a long history of heuristics, approximation algorithms, and exact algorithms for the Steiner tree problem~\cite{ahmed2019multi}.

A spanner is a subgraph that approximately preserves pairwise distances in the original graph $G$~\cite{spannersurvey}. A subset spanner needs only approximately preserve distances between a subset $T \subseteq V$ of vertices. Two common types of spanners include multiplicative $\alpha$-spanners, which preserve distances in $G$ up to a multiplicative $\alpha$ factor, and additive $+\beta$ spanners, which preserve distances up to additive $+\beta$ error. A distance preserver is a special case of the spanner where distances are preserved exactly. 
The multiplicative $\alpha$--spanner problem is NP-hard~\cite{spannersurvey}.  
Further, it is NP-hard to approximate the multiplicative $\alpha$--spanner problem to within a factor of $O(\log |V|)$, even when restricted to bipartite graphs~\cite{spannersurvey}. 
There exists a classical greedy algorithm~\cite{spannersurvey} that constructs a multiplicative $\alpha$--spanner given a graph $G$ and a real number $\alpha \geq 1$. 
It has been shown that given a weighted graph $G$ and $t \ge 1$, there is a greedy $(2t-1)$--spanner ($\alpha = 2t-1$) $H$ containing at most $n \lceil n^{1/t}\rceil$ edges, and whose weight is at most $w(MST(G)) (n/t)$ where $w(MST(G))$ denotes the weight of a minimum spanning tree of $G$. Later, this greedy spanner algorithm has been generalized for subsetwise spanners~\cite{ahmed2023multi}.

For very large graphs, additive error is arguably a much more appealing paradigm. 
It has been shown that all graphs have $+2$, $+4$, and $+6$ spanners on $O(n^{3/2})$, $O(n^{7/5})$, and $O(n^{4/3})$ edges respectively~\cite{spannersurvey}.
There are several major differences between multiplicative spanners and additive spanners. The construction of additive spanners depends on the additive error as mentioned earlier. Unlike multiplicative spanners, the number of edges in additive spanners does not always decrease as the error increases; an interesting result shows that there exists a class of graphs that do not have $+c$ spanners on $n^{4/3 - \epsilon}$ edges~\cite{spannersurvey} where $c$ and $\epsilon$ are a small integer and a positive real number respectively. Also, the algorithms for additive spanners do not naturally generalize for weighted graphs. Recently, several algorithms for weighted additive spanners have been provided that require significant changes to the algorithms of unweighted spanners~\cite{elkin2022improved,ahmed2021additive}.

\begin{figure*}[ht]
    \centering
    \includegraphics[width=2.0\columnwidth]{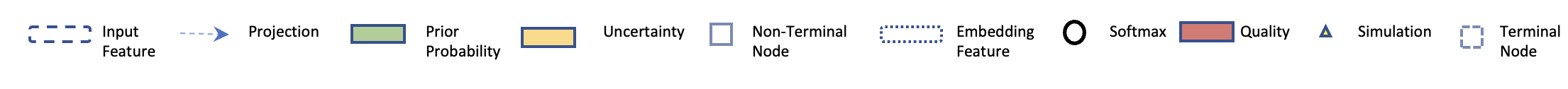}
    \label{fig:model_a}
\end{figure*}
\begin{figure*}[ht]
    \centering
    \includegraphics[width=2.0\columnwidth]{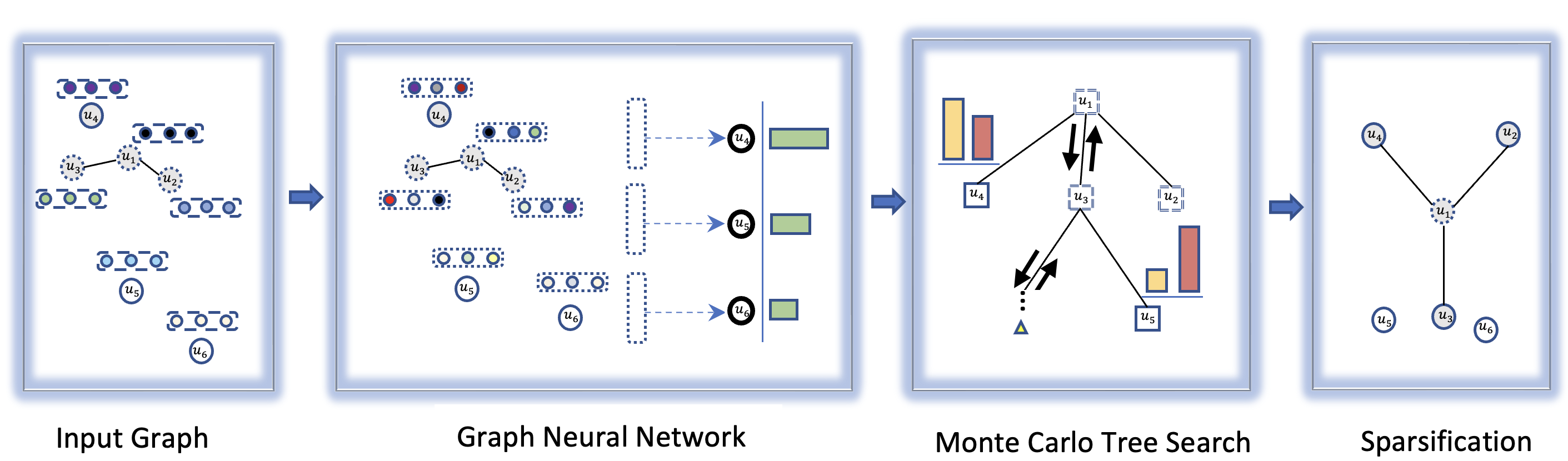}
    \caption{GNN assisted MCTS: first, train a GNN to evaluate non-terminal nodes, then use the network and heuristics to compute a Steiner tree with MCTS.}
    \label{fig:model}
\end{figure*}

\subsection{Problem Statement:}
We consider three sparsification problems: the Steiner tree, subsetwise multiplicative spanner, and subsetwise additive spanner problems.
In the Steiner Tree Problem, we are given a weighted graph $G = (V, E)$ and a set of terminals $T \subseteq V$, and we want to compute a minimum weighted subtree $H$ that spans $T$. 
The non-terminal nodes in $H$ are called the Steiner nodes. 
We denote the shortest path distance from $u$ to $v$ in the graph $G$ by $d_G(u,v)$. In the subsetwise multiplicative spanner problem, besides $G$ and $T$, we are also given a multiplicative stretch $\alpha \geq 1$, and we want to compute a subgraph $H$ such that for all $u, v \in T, d_H(u, v) \leq \alpha \cdot d_G(u, v)$. In the subsetwise additive spanner problem, instead of a multiplicative stretch $\alpha$, we are given an additive error $\beta \geq 0$. The objective is to compute a subgraph $H$ such that for all $u, v \in T, d_H(u, v) \leq d_G(u, v) + \beta W$, where $W$ is the maximum edge weight in $G$. The objective of these problems is to either minimize the total edge weights or the number of edges of $H$. 

\subsection{Summary of Contributions:}

We describe an approach for computing the above sparsifications by combining a graph neural network and Monte Carlo Tree Search (MCTS). 
We first train a graph neural network that takes as input a partial solution and proposes a new node to be added as output. This neural network is then used in an MCTS to compute a sparsification. The proposed method consistently outperforms the standard approximation algorithms on different types of graphs and often finds the optimal solution.
We illustrate our approach in Figure~\ref{fig:model}.
Our approach builds on the work of Xing et al.~\cite{xing2020graph} for TSP. Since TSP is non-trivially different from the sparsification problems, we needed to address challenges in both training the graph neural network and testing the MCTS. We summarize our contribution below:
\begin{itemize}
    \item To train the neural network we generate exact solutions of different graph sparsification instances. From each instance, we generate several data points. The purpose of the neural network is to predict a new important node, given a set of current important nodes. Since any permutation of the set of solution nodes can lead to a valid sequence of predictions, we use random permutations to generate data points for the network.
    
    \item After we select a set of important nodes for a given instance, it is not straightforward to compute a sparsification. 
    We utilize various existing algorithms to compute the sparsification from the selected set of important nodes.
    
    \item Our method can work for non-Euclidean graphs as well. We evaluate our method on some known hard instances from the SteinLib database~\cite{KMV00} that are non-Euclidean. 
    
    \item We compare our framework with different well-known approximation algorithms. 
    The experimental result shows that our method outperforms these existing algorithms. The proposed method is fully functional and available on \href{https://github.com/abureyanahmed/gnn-msts-sparsification}{GitHub}.
\end{itemize}


\section{Our approach}
\label{sec:approach}

Our approach keeps a set of important nodes $S = \{v_1, v_2, \cdots, v_i\}$ for the sparsification instance, and gradually adds more nodes in $S$. Initially, $S$ is equal to the set of terminals $T$.
Then, $\overline{S} = V - S$ is the set of candidate nodes to be added to $S$.
A natural approach is to train a neural network to predict which node to add to $S$ at a particular step. That is, neural network $f(G|S)$ takes the graph $G$ and $S$ as input, and returns probabilities for the remaining nodes, indicating the likelihood they are important for the sparsification.
We adapt the GNN of~\cite{xing2020graph} to represent $f(G|S)$.

Intuitively, we can directly use the probability values, selecting all nodes with a probability higher than a given threshold. We can then construct a subgraph from the selected nodes in different ways. For example, we can compute the induced graph of the selected nodes (add an edge if it connects to selected nodes) and extract a minimum spanning tree~\cite{cormen2009introduction} for the case of the Steiner tree problem. Note that the induced graph may be disconnected and therefore the spanning tree will be also disconnected. 
Hence it may not provide a valid solution. This issue can be addressed by reducing the given threshold until we obtain a valid solution.

However, deriving sparsifications in this fashion might not be reliable 
since it has only one chance to compute the 
solution, and it never goes back to reverse the decision. To overcome this drawback, we leverage the MCTS. In the MCTS, each tree node represents a state that is a possible set of important nodes for the sparsification problem.
We use a variant of PUCT~\cite{rosin2011multi} to balance exploration (i.e., visiting a state as suggested by the neural network policy) and exploitation (i.e., visiting a state that has the best value). The overall approach is illustrated in Figure~\ref{fig:model}. 

\subsection{Graph neural network architecture:}

\begin{figure}[ht]
\minipage{0.48\textwidth}
    \centering
    \includegraphics[width=.9\linewidth]{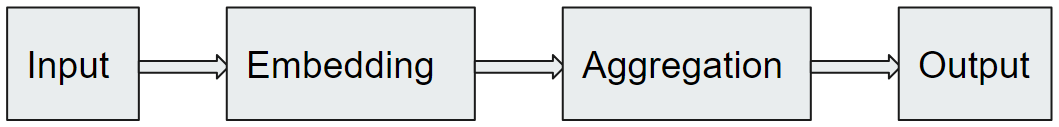}
    \subcaption{Different modules of GSE-GNN.}
    \label{fig:pipeline}
\endminipage\hfill
\minipage{0.16\textwidth}
    \centering
    \includegraphics[width=.9\linewidth]{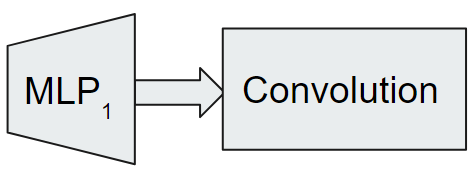}
    \subcaption{The embedding module.}
    \label{fig:embedding}
\endminipage\hfill
\minipage{0.32\textwidth}
    \centering
    \includegraphics[width=.9\linewidth]{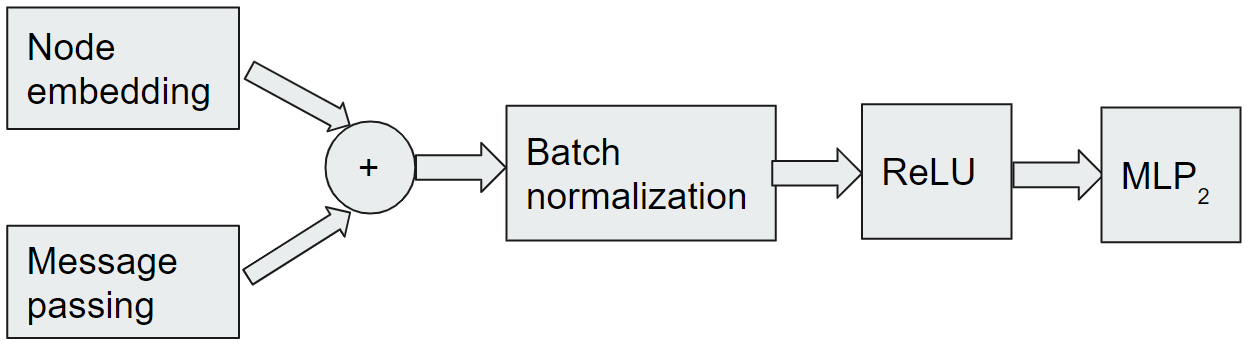}
    \subcaption{The convolution module.}
    \label{fig:convolution}
\endminipage\hfill
\caption{The generalized static edge graph neural network (GSE-GNN) model.}
\label{fig:GSE-GNN}
\end{figure}

Some combinatorial problems like the independent set problem and minimum vertex cover problem do not consider edge weights. However, edge weight is an important feature of the sparsification problem as the objective and shortest path distances are computed based on the weights. Hence, we adapt the static edge graph neural network (SE-GNN)~\cite{xing2020graph}, to efficiently extract node and edge features. The SE-GNN model only works for Euclidean graphs due to the dependency of node positions. Our generalized SE-GNN (GSE-GNN) model can handle non-Euclidean graphs as well. We illustrate the architecture of the GSE-GNN model in Figure~\ref{fig:GSE-GNN}.

\subsubsection{The input module:}

To train a neural network, information about the structures of the concerned graph, terminal nodes, and contextual information, i.e., the set of important nodes $S$, is required. We tag node $u$ with $x^t_u=1$ if it is a terminal, otherwise $x^t_u=0$. We also tag $u$ with $x^a_u=1$ if it is already added, otherwise $x^a_u=0$.
The SE-GNN model only considers Euclidean graphs since it tags each node by the position of the nodes. For non-Euclidean graphs, there is no trivial way to compute the coordinates of the nodes. In our GSE-GNN model, we resolve this issue by computing the coordinates of non-Euclidean graphs using a spring embedder~\cite{kobourov2012spring}. Besides that, we also tag each node by several other properties of the input instance: node degree, clustering coefficients~\cite{saramaki2007generalizations}, and different node centrality values~\cite{bonacich1987power}. Let $x_u$ be the feature vector containing all the tags of node $u$.
Intuitively, $f(G|S)$ should summarize the state of such a ``tagged'' graph (a concatenation of all the feature vectors) and generate the prior probability for each node to get included in $S$.

\subsubsection{The embedding module:}

The embedding module has a multi-layer perceptron MLP$_1$ that maps a feature vector $x_u$ of node $u$ to a higher embedding space vector $H_u^0$, see Figure~\ref{fig:embedding}. The multi-layer perceptron MLP$_1$ is followed by a convolution module that consists of a stack of $L$ neural network layers, where each layer aggregates local neighborhood information, i.e., features of neighbors of each node, and then passes this aggregated information to the next layer. This procedure of aggregating neighborhood information is known as message-passing; the original SE-GNN model only considered the graph convolutional message-passing procedure~\cite{kipf2016semi} whereas we incorporate the graph attention procedure~\cite{velivckovic2017graph} as well.
We use $H_u^l \in \mathbb{R}^d$ to denote the real-valued feature vector associated with node $u$ at layer $l$. Specifically, the basic GNN model~\cite{hamilton2017inductive} can be implemented as follows. In layer $l = 1, 2, \cdots, L$, a new feature is computed as given by~\ref{eqn:gnn}.

\begin{equation}\label{eqn:gnn}
H_u^{l+1} = \sigma \Big( \theta_1^l H_u^l + \sum_{v \in N(u)} \theta_2^l H_v^l \Big)
\end{equation}

In~\ref{eqn:gnn}, $N(u)$ is the set of neighbors of node $u$, $\theta_1^l$, and $\theta_2^l$ are the parameter matrices for layer $l$, and $\sigma(\cdot)$ denotes a component-wise non-linear function such as a ReLU function. 
The edge features are 
not taken into account in~\ref{eqn:gnn}. There are some edge features, e.g. edge weights and common neighborhoods~\cite{liben2003link}, that we want to incorporate into our model. We denote the edge features of edge $uv$ by $e_{uv}$.
Some previous methods~\cite{xie2018crystal} use the following equation to incorporate edge features.

\begin{equation}\label{eqn:se-gnn}
H_u^{l+1} = \sigma \Big( \theta_1 x_u + \theta_2 \sum_{v \in N(u)} H_v^l + \theta_3 \sum_{v \in N(u)} \sigma(\theta_4 e_{uv}) \Big)
\end{equation}


In~\ref{eqn:se-gnn}, $\theta_1$, $\theta_2, \theta_3$, and $\theta_4$ are all model parameters. We can see in~\ref{eqn:gnn} and~\ref{eqn:se-gnn} that the nonlinear mapping of the
aggregated information is a single-layer perceptron, which is not enough to map distinct multisets into unique embeddings. Hence, as suggested in~\cite{xing2020graph}, we replace the single perceptron with a multi-layer perceptron. Finally, we compute a new node feature $H_u$ using~\ref{eqn:se-gnn2}.

\begin{equation}\label{eqn:se-gnn2}
H_u^{l+1} = \text{MLP}_2^l \Big( \theta_1^l H_u^l + \sum_{v \in N(u)} \theta_2^l H_v^l + \sum_{v \in N(u)} \theta_3^l e_{uv} \Big)
\end{equation}

In~\ref{eqn:se-gnn2}, $\theta_1^l$, $\theta_2^l$, and $\theta_3^l$ are parameter matrices, and MLP$_2^l$ is the multi-layer perceptron for  layer $l$. 


\subsubsection{The aggregation and output modules:}

Once the feature for every node is computed after updating $L$ layers, we aggregate the new feature vector by summing up all the elements of the vector. We then pass that aggregated value to the softmax function ($\text{softmax}(z) = e^z/\sum_{i}e^{z_i}$) and denote it by $f(G|S; \theta)$. This function $f(G|S; \theta)$ returns the prior probability for each node indicating how likely is the node to be in $S$. Specifically, we fuse all node feature $H_u^L$ as the current state representation of the graph and parameterize $f(G|S; \theta)$ as expressed by~\ref{eqn:se-gnn-out}.

\begin{equation}\label{eqn:se-gnn-out}
f(G|S; \theta) = \text{softmax}( \text{sum}(H_1^L) , \cdots , \text{sum}(H_{|V|}^L) ) 
\end{equation}

Here, $\text{sum}(z) = \sum_i z_i$. During training, we minimize the cross-entropy loss for each training sample $(G_i , S_i )$ in a supervised manner as given by~\ref{eqn:se-gnn-loss}.

\begin{equation}\label{eqn:se-gnn-loss}
\ell(S_i, f(G_i|S_i; \theta)) = - \sum_{j=|T_i|+1}^N y_j^T \log f(G_i|S_i(1:j-1); \theta)  
\end{equation}

In~\ref{eqn:se-gnn-loss}, $S_i$ is an ordered set of important nodes of a sparsification which is a permutation of a subset of the nodes of graph $G_i$, with $S_i(1:j-1)$ the ordered subset containing the first $j-1$ elements of $S_i$, $N$ the number of nodes in the sparsification, $y_j^T$ the transpose of $y_j$, and $y_j$ a 
vector of length $|V|$ with 1 in the $S_i(j)$-th position and 0 otherwise. We provide more details in Section~\ref{sec:training}.

\subsection{GNN assisted MCTS:}

Several recent GNN-based models for solving combinatorial problems leverage different kinds of greedy or tree searches~\cite{dai2017learning,li2018combinatorial,xing2020graph}.
We use an MCTS for our sparsification problems.
The search space of a sparsification instance can be huge. In a traditional MCTS, random sampling of the search space gradually expands the search tree. 
Our graph neural network assisted MCTS (GNN-MCTS) adds new nodes in the search tree from the prediction of GSE-GNN instead of random sampling. 

For each MCTS node $v$, there is an action space $A(v)$. Each action $a \in A(v)$ represents a node in $\overline{S}$. 
The MCTS counts the number of times a particular action $a$ has been selected from an MCTS node $v$ to compute the uncertainty of $a$ from $v$. We denote this action count by $N(v, a)$. We adapt the standard PUCT~\cite{rosin2011multi} algorithm to compute the uncertainty $U(v,a)$ of $a$ from $v$. Similar to PUCT~\cite{rosin2011multi}, we set the value of $U(v,a)$ equal to $c_{puct}P(v,a)\frac{\sqrt{\sum_b N(v, b)}}{1+N(v, a)}$ where $c_{puct}$ is a tuning parameter and $P(v,a)$ is the neural network policy. 

Another important quantity of our MCTS is the quality $Q(v,a)$ of an action $a$ from an MCTS node $v$. 
Let $H$ be the sparsification after executing $a$ from $v$. We denote the cost of $H$ by $\text{cost}(H)$. Notice that the value $\text{cost}(H)$ can be large if the number of edges in $H$ is relatively larger. However, the standard MCTS takes quality values in the range $[0, 1]$~\cite{rosin2011multi}.
We address this issue by normalizing the sparsification cost $\text{cost}(H)$ as suggested by~\cite{xing2020graph}. We set the quality $Q(v,a)$ equal to $\frac{\text{cost}(H) - w}{b-w}$ where $b$ and $w$ are the minimum and maximum sparsification costs among the actions of $v$.

Following the standard MCTS algorithm, we aggregate the quality and uncertainty; and select the best action according to the aggregated value. We also strengthen the MCTS by selecting an action uniformly with a small probability $\epsilon$ to better explore the search space~\cite{sutton2018reinforcement}. In other words, at time step $t$, our MCTS selects the action $a_t$ with probability $1-\epsilon$ such that:
\begin{equation}\label{eqn:mcts}
a_t = \text{arg} \text{max}_a(Q(v_t, a) + U(v_t, a))
\end{equation}
And with a small probability $\epsilon$, the MCTS selects randomly from among all the nodes in $\overline{S}$ with equal probability.
Each round of the MCTS consists of four steps:
\begin{itemize}
    \item Selection: The MCTS selects a leaf node $u$ starting from the root node using~\ref{eqn:mcts}.
    \item Expansion: The MCTS creates a new leaf node $v$ such that $v$ is the child of selected node $u$.
    \item Simulation: The MCTS gradually adds nodes from $\overline{S}$ using the neural network prediction. After each addition, the MCTS computes a sparsification (as described later). The number of nodes added is the sample size. Finally, the MCTS selects the best sample and updates the state of $v$ accordingly.
    \item Backpropagation: The MCTS updates the best and worst costs from the state of $v$ to its ancestors.
\end{itemize}

Our MCTS is similar to a recent MCTS proposed for computing TSP~\cite{xing2020graph}. However, there are several major changes in our method as described below:
\begin{itemize}
    \item The graph sparsification problem is significantly different from TSP that was considered in~\cite{xing2020graph}. Unlike the sparsification problem, all nodes must be present in a traveling salesman tour. Hence in the MCTS of~\cite{xing2020graph}, initially the set $S$ was empty, and gradually they added all the nodes in $S$. However, in the sparsification problem, all terminals must be in the final solution. Hence at the beginning of our search, the set $S$ contains all terminals. Our initial experiment also showed that starting with a set $S$ that contains all terminals significantly improves the running time than starting from an empty set.
    \item 
    The sample size of TSP is huge since different permutations of the nodes provide different tours. 
    A sparsification is the same for different permutations and a large sample size will increase the running time as well since we compute a shortest path tree for each additional node in $O(n\log n)$ time~\cite{cormen2009introduction}. Hence we keep the sample size relatively small.
    Details are provided in the following sections. 
    \item 
    Since we keep the sample size relatively small, we strengthen the exploration process by mixing in a random search strategy that has been found effective in reinforcement learning~\cite{sutton2018reinforcement}: use the uncertainty value from the count of visited nodes most of the time, but every once in a while, say with small probability $\epsilon$, select randomly from among all the nodes in $\overline{S}$ with equal probability.
\end{itemize}

\begin{figure}[ht]
    \centering
    \includegraphics[width=.6\linewidth]{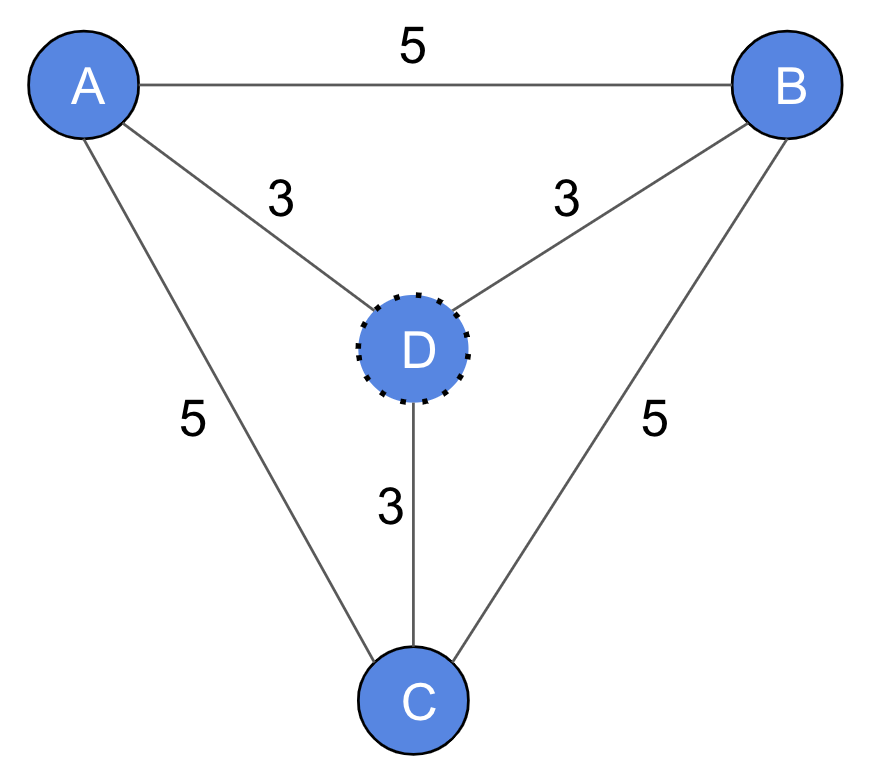}
    \caption{Example graph for the Steiner tree heuristic. Considering $D$ as a terminal node and computing the MST on the metric closure provides a better solution than the 2-approximation algorithm.}
    \label{fig:heuristic}
\end{figure}

\subsection{Computing a sparsification from $S$:\label{sec:heuristics}}
Our heuristics are motivated by existing algorithms of sparsification problems. An algorithm for a sparsification problem takes the set of terminals $T$ as a parameter. Our MCTS uses the same algorithm, however, instead of $T$, the MCTS uses $S$ as the set of terminals. Initially, the MCTS sets $S = T$ as described above and gradually adds more nodes using the guidance of the GNN. After computing the sparsification, the MCTS applies different pruning algorithms since $S$ usually contains more nodes than $T$.
We now describe the existing algorithms we have used.
\begin{enumerate}
    \item \textbf{The 2-approximation algorithm for computing Steiner trees:} In this algorithm~\cite{agrawal1995trees}, given an input graph $G = (V, E)$ and a set of terminals $S$, we first compute a metric closure graph $G' = (S, E')$. Every pair of nodes in $G'$ is connected by an edge with a weight equal to the shortest path distance between them in $G$.
    The minimum spanning tree of the metric closure provides a 2-approximation to the optimal Steiner tree (if $S=T$). 
    The MCTS improves the quality by adding new nodes in $S$.
    For example, in Figure~\ref{fig:heuristic}, $A$, $B$, and $C$ are terminal nodes and $D$ is not. 
    Note that $D$ does not appear in any shortest path as each shortest path distance between pairs of terminals is 5 and none of them goes through $D$. 
    Without loss of generality, the $2$-approximation algorithm (when $S=T$) chooses the $A-C-B$ path with a total cost of 10, while the optimal solution that uses $D$ has a cost of 9. 
    While the $2$-approximation algorithm (when $S=T$) does not consider any node that does not belong to a shortest path between two terminal nodes, the MCTS considers such nodes. 
    \item \textbf{The greedy algorithm for computing subsetwise multiplicative spanners:} In this greedy algorithm~\cite{ahmed2023multi}, we are also given a multiplicative stretch $\alpha$. We again first compute a metric closure graph. Then we sort the edges of the metric closure in non-decreasing order of weights. Initially, the sparsification $H$ does not contain any edges. We go through each edge $e = uv$ according to the sorted order and add it in $H$ if $\alpha \cdot w(e) \leq dist_H(u, v)$. 
    Finally, we replace each abstract edge of $H$ with the corresponding shortest path of $G$.
    \item \textbf{The subsetwise $+2W$ algorithm for computing additive spanners:} 
    Here, the additive stretch $\beta=2W$, where $W$ is the maximum edge weight of $G$. There exist several algorithms for this problem; a recent study compares different algorithms~\cite{ahmed2021multi}. We use an algorithm in this paper that performs well in practice. This algorithm starts with an empty set $H$ and for each node in $G$, it adds $|S|^{2/3}$ lightest neighboring edges in $H$. Later, it adds some more edges to $H$ such that for all $u, v \in S, dist_H(u,v) \leq dist_G(u, v) + 2W$. We call this algorithm the subsetwise $+2W$ algorithm.
\end{enumerate}
Since the MCTS adds additional nodes in $S$, at the end of the algorithm we prune some nodes and edges that are not necessary. We now describe the pruning algorithms that we have used.
\begin{enumerate}
    \item \textbf{Pruning for Steiner trees:} Let $H$ be the output of the 2-approximation. Since our goal is to compute a tree, we remove some edges from $H$ if there exist any cycles. To do that, we compute a minimum spanning tree $H'$ of $H$. A node is a pendant node if it has a degree equal to one. We then check whether there exist any pendant nodes that are not in $T$. We remove all pendant nodes not in $T$ from $H'$. We denote the new tree by $H''$. We return $H''$ as the final output.
    \item \textbf{Pruning for spanners:} We sort all the edges of the computed spanner $H$ in the decreasing order of edge weights. We go through each edge $e$ in this order and delete $e$ from $H$ if $H-e$ is a valid spanner. Note that, we use the same pruning algorithm for multiplicative and additive spanners.
\end{enumerate}

\section{Model setup and training}\label{sec:training}

Our training data consists of input graphs $G = (V, E)$, edge weights $w:E \rightarrow \mathbb{R}^+$, terminals $T \subseteq V$, and a stretch value depending on the type of sparsification. 
Given $G, w, T$, and a stretch value (for spanner instances), our goal is to give label 1 to the next node to be added and 0 to all others.
Initially, we set $S=T$ as all terminals must be in the sparsification. Consider a graph with 6 nodes $u_1, u_2, \cdots, u_6$, a set of terminals $T=\{u_1, u_2, u_3\}$, and an optimal sparsification $H$ contains the first five nodes $u_1, u_2, \cdots, u_5$. For this example, initially, we set $S = T = \{u_1, u_2, u_3\}$. Since we have two non-terminal nodes $u_4$ and $u_5$ in $H$, both permutations $u_4, u_5$ and $u_5, u_4$ are valid. For the first permutation, after setting $S=\{u_1, u_2, u_3\}$, the next node to be added to the solution is $u_4$. Hence for this data point, only the label for $u_4$ is 1. This permutation provides another data point where $S = \{u_1, u_2, u_3, u_4\}$ and only the label for $u_5$ is equal to 1. Similarly, we can generate two more data points from the other permutation. This exhaustive consideration of all possible permutations does not scale to larger graphs, so we randomly select at most 100 permutations from each optimal solution. 
The model is trained using the ADAM optimizer~\cite{kingma2014adam} to minimize the cross-entropy loss between the model's prediction and the ground truth (a vector in $\{0, 1\}^{|V|}$ indicating whether a node is the next solution node or not) for each training sample.

\subsection{Data generation:}
We produce sparsification instances using the random geometric graph generation model~\cite{penrose2003random}. 
Let $n$ be the number of nodes of the graph.
In the random geometric graph model, we uniformly select $n$ points from the Euclidean cube, and connect nodes whose Euclidean distance is not larger than a threshold $r$. If $r \geq \sqrt{\frac{\ln n}{\pi n}}$, then the graph is connected with high probability. To produce relatively denser graphs, we set $r = \sqrt{\frac{2\ln n}{\pi n}}$.

We generate Steiner tree, multiplicative, and additive spanner instances using the above random graph generation model. We assign random integer weights in the range $\{1, 2, \cdots, 10\}$ to each edge. As discussed earlier, we set the multiplicative stretch $\alpha=2$ and the additive stretch $\beta = 2W$, where $W$ is the maximum edge weight of the graph.
The number of nodes is in $\{20, 50, 100\}$. We randomly select half of the nodes of each graph and set them as terminals.

For the Steiner tree and multiplicative spanner problems, we train the graph neural network on 5000 random geometric instances of 100 nodes. For the additive spanner problem, we train on the same number of geometric instances of 50 nodes. We use smaller instances for additive spanners because it is not possible to compute optimal solutions of larger instances by the maximum 20 hours time limit we use. Each of these instances generates multiple training data points from different permutations of non-terminal nodes as described above. The number of nodes in the test dataset of MCTS is in $\{20, 50, 100\}$.
As random geometric instances can be ``easy" to solve, we also evaluate our approach on graphs from the SteinLib library~\cite{KMV00}, which provides hard instances for the Steiner tree problem.
Specifically, we perform experiments on two SteinLib datasets: \href{http://steinlib.zib.de/showset.php?I080}{I080} and \href{http://steinlib.zib.de/showset.php?I160}{I160}.
Each instance of the I080 and I160 datasets contains 80 nodes and 160 nodes respectively. 
Unlike geometric graphs, these datasets contain non-Euclidean graphs. We use the spring embedder~\cite{kobourov2012spring} to compute the positions of SteinLib instances as one of our input features is node position.


\begin{figure*}[ht]
\minipage{0.32\textwidth}
  \includegraphics[width=\linewidth]{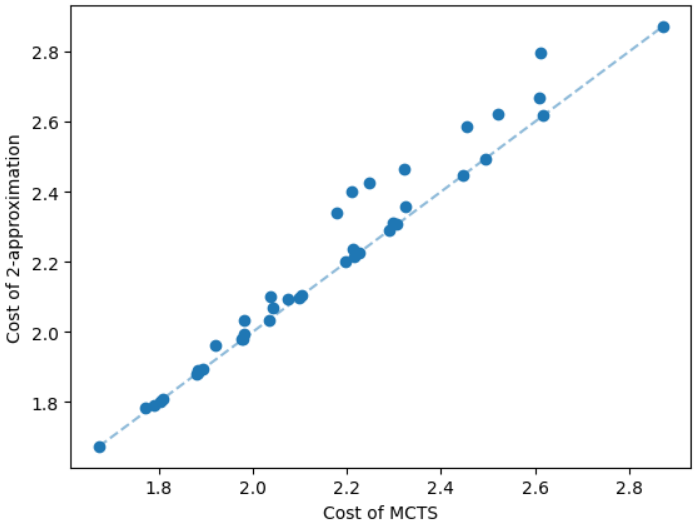}
  \subcaption{Twenty nodes graphs}
  \label{fig:GE20_GNN_APX}
\endminipage\hfill
\minipage{0.32\textwidth}
  \includegraphics[width=\linewidth]{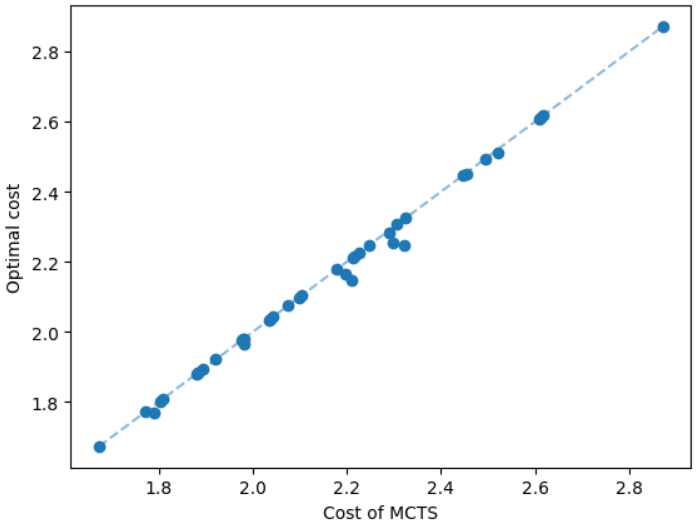}
  \subcaption{Twenty nodes graphs}
  \label{fig:GE20_GNN_OPT}
\endminipage\hfill
\minipage{0.32\textwidth}
  \includegraphics[width=\linewidth]{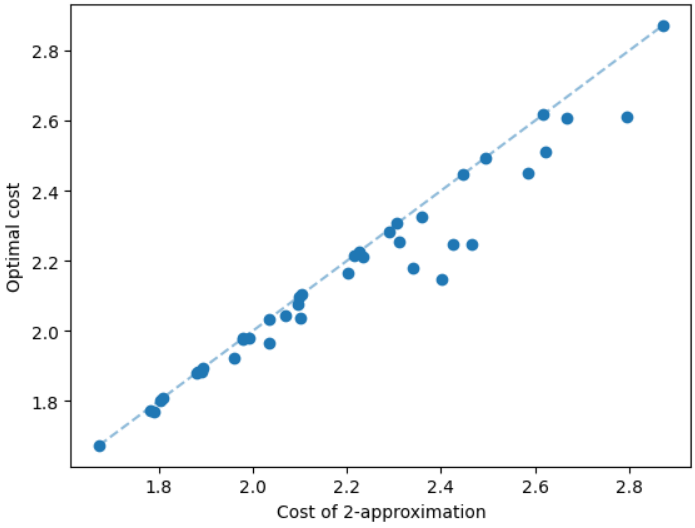}
  \subcaption{Twenty nodes graphs}
  \label{fig:GE20_APX_OPT}
\endminipage\hfill

\minipage{0.32\textwidth}
  \includegraphics[width=\linewidth]{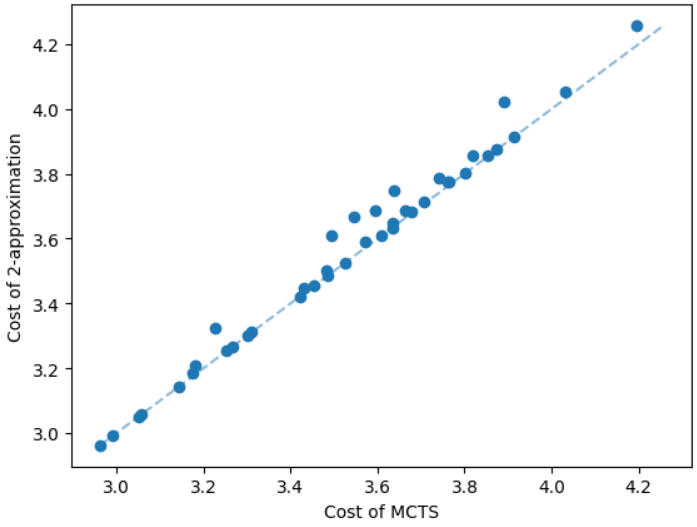}
  \subcaption{Fifty nodes graphs}
  \label{fig:GE50_GNN_APX}
\endminipage\hfill
\minipage{0.32\textwidth}
  \includegraphics[width=\linewidth]{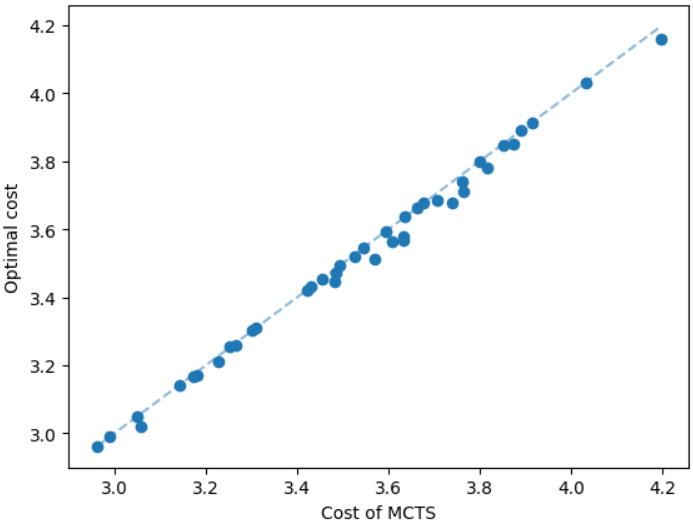}
  \subcaption{Fifty nodes graphs}
  \label{fig:GE50_GNN_OPT}
\endminipage\hfill
\minipage{0.32\textwidth}
  \includegraphics[width=\linewidth]{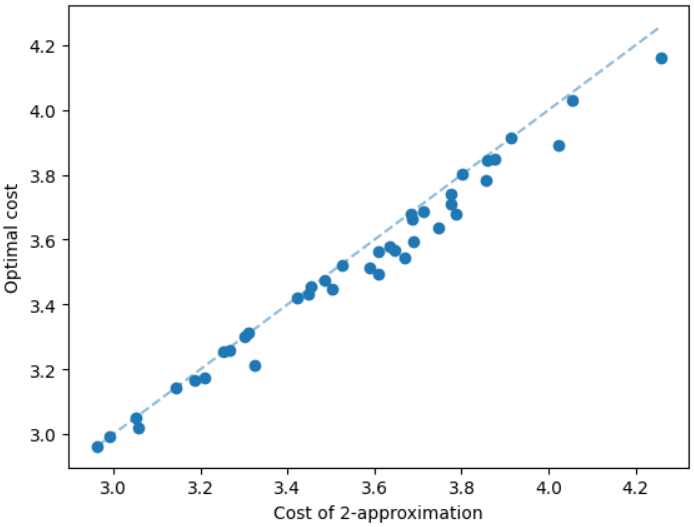}
  \subcaption{Fifty nodes graphs}
  \label{fig:GE50_APX_OPT}
\endminipage\hfill

\minipage{0.32\textwidth}%
  \includegraphics[width=\linewidth]{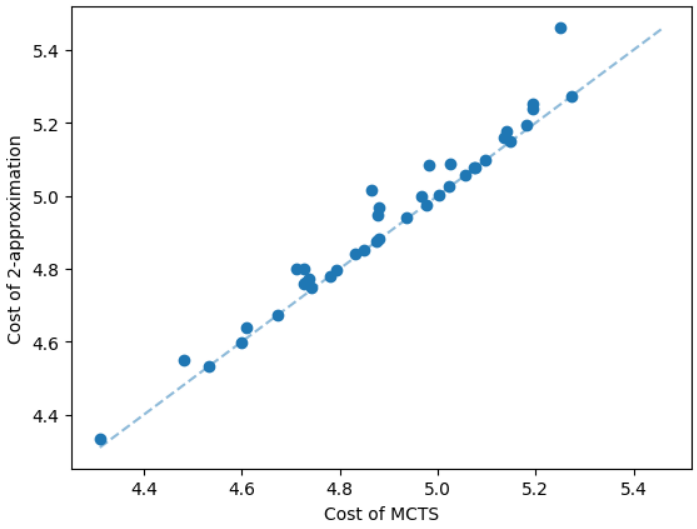}
  \subcaption{Hundred nodes graphs}
  \label{fig:GE100_GNN_APX}
\endminipage\hfill
\minipage{0.32\textwidth}%
  \includegraphics[width=\linewidth]{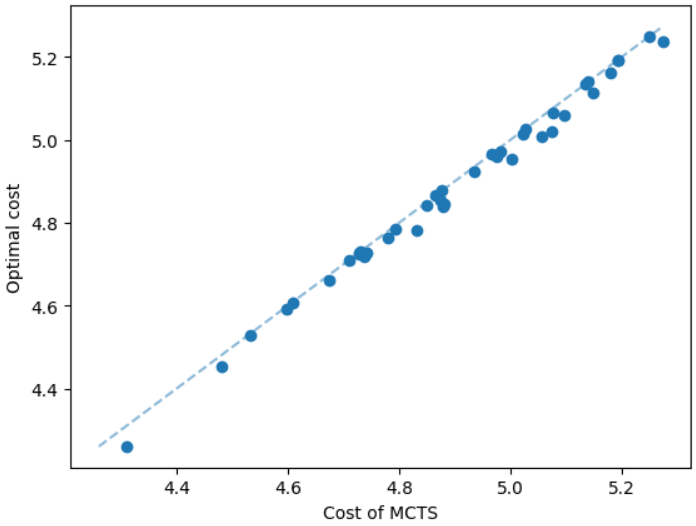}
  \subcaption{Hundred nodes graphs}
  \label{fig:GE100_GNN_OPT}
\endminipage\hfill
\minipage{0.32\textwidth}%
  \includegraphics[width=\linewidth]{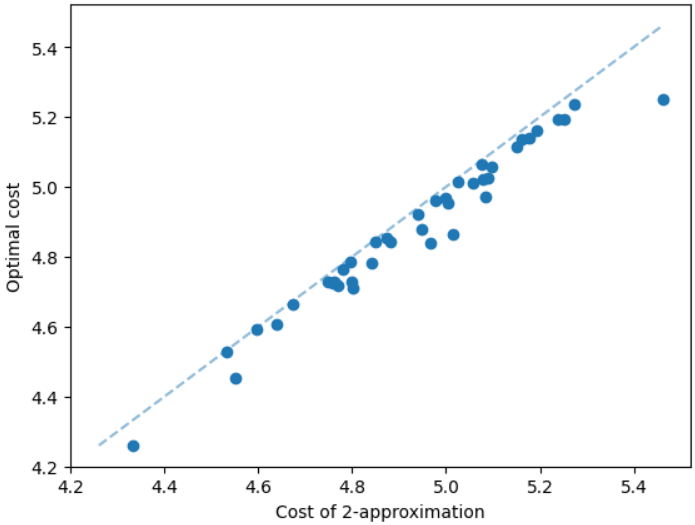}
  \subcaption{Hundred nodes graphs}
  \label{fig:GE100_APX_OPT}
\endminipage\hfill
\caption{Performance on random geometric Steiner instances. The lower the cost the better the algorithm is. Our algorithm (MCTS) is nearly optimal and performs better than 2-approximation.}
\label{fig:performance_GE_ST}
\end{figure*}

\subsection{Computing optimal solutions:}
We need to compute the optimal solutions to evaluate the performance of our approach (and other existing algorithms).
There are different integer linear programming (ILP) models for the sparsification problems. The cut-based approach considers all possible combinations of partitions of terminals and ensures that there is an edge between that partition. This ILP is simple but introduces an exponential number of constraints. 
A better ILP approach in practice considers a set of terminals as source nodes and sends a flow to the rest of the terminals; see~\cite{ahmed2019multi,kobourovapproximation} for details about these and other ILP methods for the exact sparsification problems.

We compute the exact solution with the flow-based ILP. We use CPLEX 20.10 as the ILP solver on a high-performance computer (Lenovo NeXtScale nx360 M5 system with 400 nodes with 192 GB of memory each). We use Python 3.10 to implement the algorithms described above.

\begin{figure*}[ht]
\minipage{0.32\textwidth}
  \includegraphics[width=\linewidth]{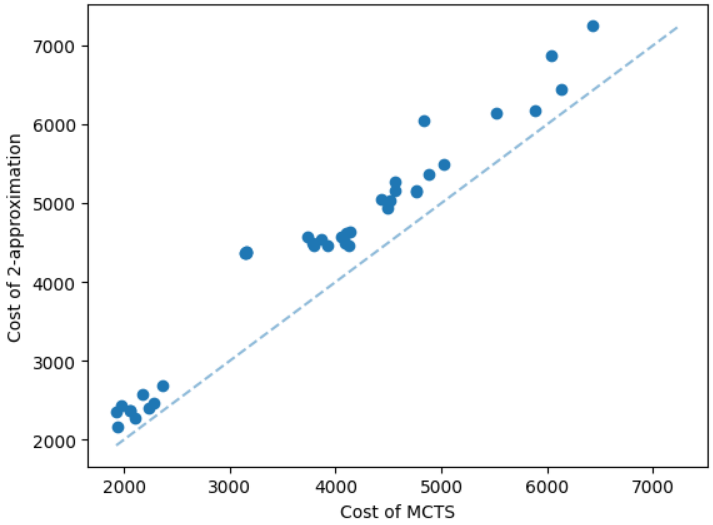}
  \label{fig:I80_GNN_APX}
\endminipage\hfill
\minipage{0.32\textwidth}
  \includegraphics[width=\linewidth]{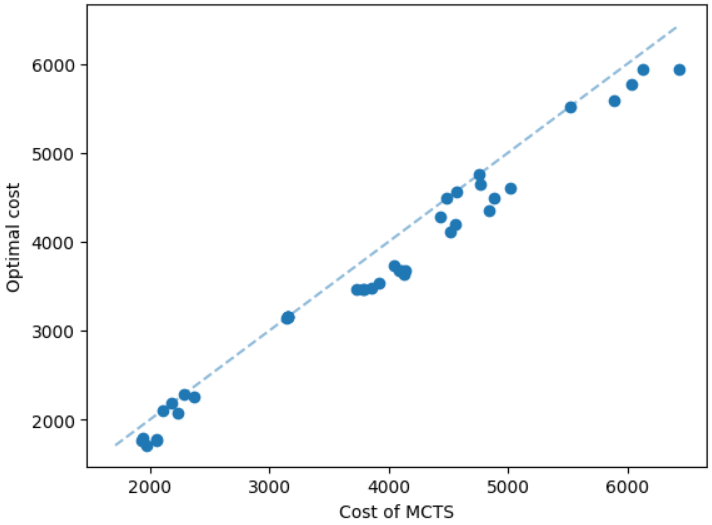}
  \label{fig:I80_GNN_OPT}
\endminipage\hfill
\minipage{0.32\textwidth}
  \includegraphics[width=\linewidth]{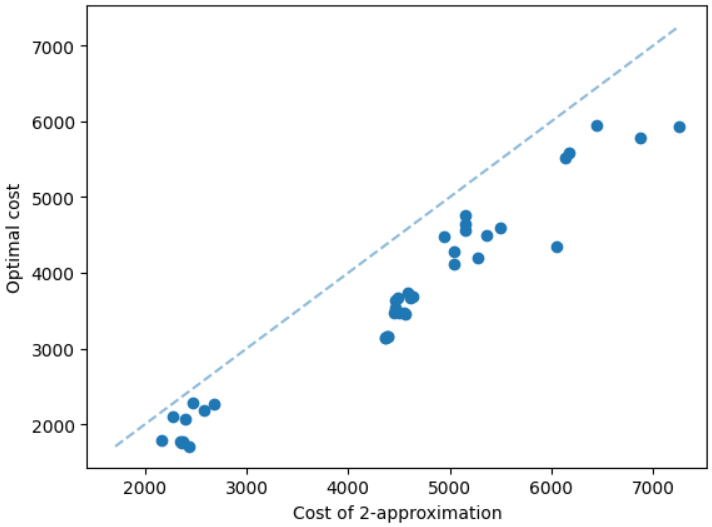}
  \label{fig:I80_APX_OPT}
\endminipage\hfill

\caption{Performance on SteinLib I080 dataset. The lower the cost the better the algorithm is. Our algorithm (MCTS) performs better than 2-approximation.}
\label{fig:I80}
\end{figure*}

\subsection{GNN architecture:}
We illustrate the architecture of our GNN in Figure~\ref{fig:GSE-GNN}. We use a 12-dimensional node feature vector that includes node positions, an indicator for terminal nodes, an indicator for solution nodes, node degree, clustering coefficients~\cite{saramaki2007generalizations}, and different node centrality values~\cite{bonacich1987power}. For edge features, we use the edge weight and common neighborhoods~\cite{liben2003link}. The input feature vector is embedded into a higher dimension using a multi-layer perceptron (MLP). We keep three hidden layers and use ReLU activation in the MLP. We set the embedding dimension equal to 128.
We use a graph convolutional network (GCN)~\cite{kipf2016semi} as a message-passing procedure for our experimental analysis and provide more details about this design choice in the Appendix.

As discussed in Section~\ref{sec:approach}, we use another MLP before mapping the node embedding into the probability space. We use two hidden layers and ReLU activation for that MLP. We have noticed that the GNN achieves good accuracy after 30 epochs and gets saturated during training. Hence we set the maximum epoch equal to 60 with early stopping equal to 15 (the model will automatically stop training when the chosen metric does not improve for 15 epochs).


\subsection{MCTS parameters:}
We set $c_{puct}=1.3$ according to our initial experiment as well as following the suggestions from previous experimental results~\cite{xing2020graph}. 
With $\epsilon$ probability, the MCTS selects an action uniformly. We set $\epsilon=0.1$ since that gives us a reasonable performance confirming existing literature~\cite{sutton2018reinforcement}. The MCTS gradually adds new nodes in the simulation step. We set the number of new nodes added at most $n$ where $n$ is the number of nodes in the input graph. We stop the MCTS when the height of the search tree is equal to $20\%$ of the number of nodes in the input graph. We discuss the reason for these design choices in the Appendix.

\begin{figure*}[ht]
\minipage{0.32\textwidth}
  \includegraphics[width=\linewidth]{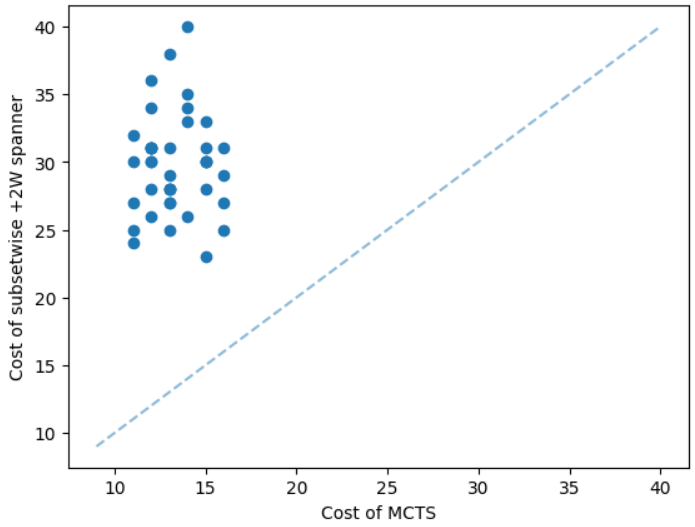}
  \subcaption{Twenty nodes graphs}
  \label{fig:GE20_GNN_APX_additive}
\endminipage\hfill
\minipage{0.32\textwidth}
  \includegraphics[width=\linewidth]{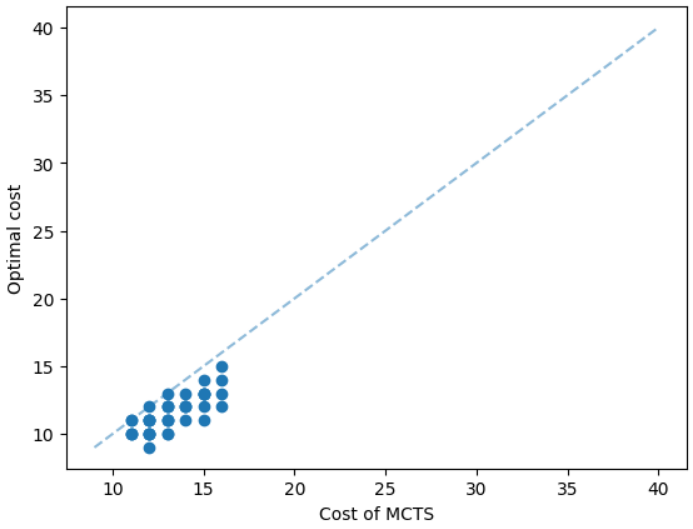}
  \subcaption{Twenty nodes graphs}
  \label{fig:GE20_GNN_OPT_additive}
\endminipage\hfill
\minipage{0.32\textwidth}
  \includegraphics[width=\linewidth]{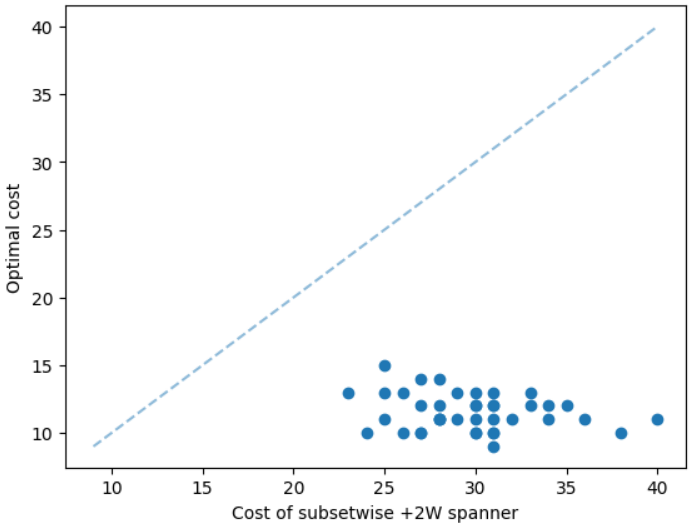}
  \subcaption{Twenty nodes graphs}
  \label{fig:GE20_APX_OPT_additive}
\endminipage\hfill

\minipage{0.32\textwidth}
  \includegraphics[width=\linewidth]{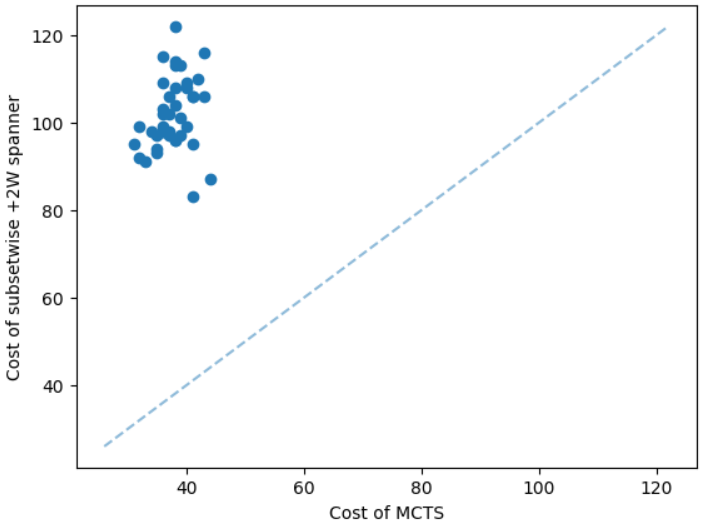}
  \subcaption{Fifty nodes graphs}
  \label{fig:GE50_GNN_APX_additive}
\endminipage\hfill
\minipage{0.32\textwidth}
  \includegraphics[width=\linewidth]{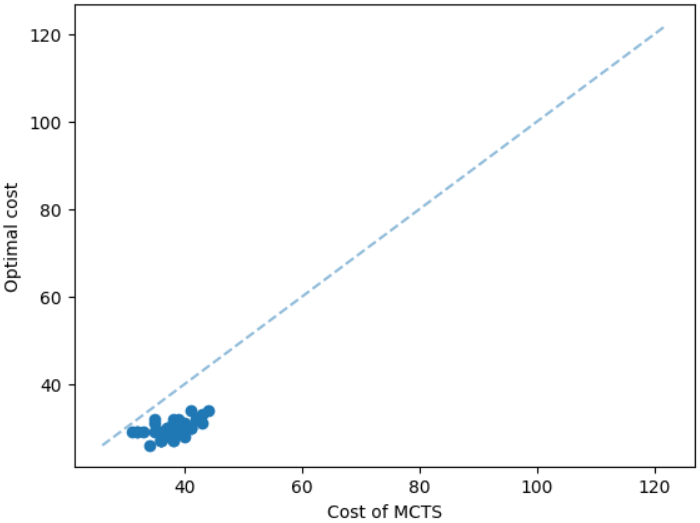}
  \subcaption{Fifty nodes graphs}
  \label{fig:GE50_GNN_OPT_additive}
\endminipage\hfill
\minipage{0.32\textwidth}
  \includegraphics[width=\linewidth]{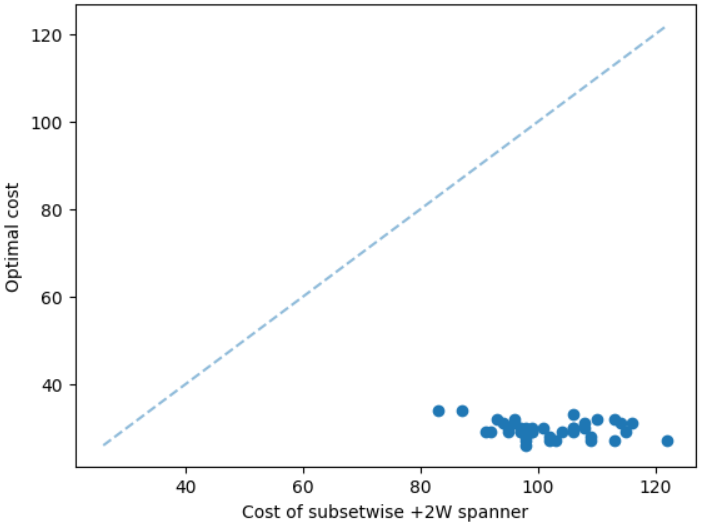}
  \subcaption{Fifty nodes graphs}
  \label{fig:GE50_APX_OPT_additive}
\endminipage\hfill

\caption{Performance on random geometric additive spanner instances ($\beta = 2W$). The lower the cost the better the algorithm is. Our algorithm (MCTS) is nearly optimal and performs better than the subsetwise +2W spanner.}
\label{fig:performance_GE_additive}
\end{figure*}

\section{Experimental results}\label{sec:experiment}

We evaluate the performance of the proposed approach by comparing the computed sparsification to those computed by the standard algorithms described in Section~\ref{sec:heuristics} and the optimal solutions. The proposed approach never performs worse than the standard algorithms. We also report running times. 

The results for geometric graphs on the Steiner tree problem are shown in Figure~\ref{fig:performance_GE_ST}. We train the model only on geometric graphs having 100 nodes. We test the MCTS on geometric graphs of different node sizes. We illustrate the performance of different algorithms on geometric graphs having 20 nodes in the top row of Figure~\ref{fig:performance_GE_ST}. We illustrate a comparison between the MCTS and the standard 2-approximation algorithm in Figure~\ref{fig:GE20_GNN_APX}. 
We can see that the costs of the MCTS are noticeably smaller compared to the 2-approximation algorithm for several instances. As illustrated in Figure~\ref{fig:GE20_GNN_OPT}, the cost difference between the MCTS and the optimal solution is significantly smaller. On the other hand, the cost of 2-approximation is relatively larger compared to the optimal cost as illustrated in Figure~\ref{fig:GE20_APX_OPT}. We illustrate the performance of different algorithms on geometric graphs having 50 nodes and 100 nodes in the middle and bottom rows of Figure~\ref{fig:performance_GE_ST} respectively.

It is natural that our method will perform well on geometric graphs since it has been trained on geometric graphs as well. A more interesting experiment would be to run our method on graphs not generated from the same generator. Not only the graphs are not geometric, but also these graphs are from the SteinLib library that contains different datasets of hard Steiner tree instances. We test our MCTS algorithm on the I080 SteinLib dataset; each of these instances contains 80 nodes and six of these nodes are terminals. We illustrate a cost comparison of this dataset in Figure~\ref{fig:I80}.

We discuss the experimental results of the subsetwise multiplicative spanner problem in the Appendix due to the page limit. We here discuss the experimental results of the subsetwise additive spanner problem. For this problem, we consider geometric graphs as before, however, we train the GNN on instances having 50 nodes instead of 100 nodes. The additive spanner problem is relatively harder~\cite{spannersurvey} and computing an exact solution also takes significantly more running time. We set a time limit equal to 20 hours to compute an exact solution, and additive spanner instances having 100 nodes need more than the limit. We test the MCTS on instances having 20 and 50 nodes. Here, we compare our method with the subsetwise $+2W$ algorithm that performs well in practice~\cite{ahmed2021multi}. The results are illustrated in Figure~\ref{fig:performance_GE_additive}. We can see that the MCTS performs significantly well compared to the subsetwise $+2W$ algorithm and generates nearly optimal solutions.

\begin{table}[ht]
\tiny
    \begin{center}
    \begin{tabular}{|p{0.7cm}|p{.4cm}|p{.4cm}|p{.4cm}|p{.4cm}|p{.4cm}|p{.4cm}|p{.5cm}|p{.4cm}|p{.5cm}|}
    \hline
     Graphs/ Algo. & ST 20 & ST 50 & ST 80 & ST 100 & MS 20 & MS 50 & MS 100 & AS 20 & AS 50 \\ 
     \hline
     Approx. & 0.16 & 0.74 & 1.29 & 1.86 & 0.21 & 2.38 & 10.92 & 0.27 & 2.72 \\  
     \hline
     MCTS & 0.64 & 3.90 & 5.77 & 6.32 & 0.98 & 9.83 & 57.17 & 1.24 & 11.79 \\  
     \hline
     OPT & 5.92 & 165.8 & 1051 & 3139 & 11.79 & 318.9 & 16139 & 37.19 & 19107 \\  
     \hline
    \end{tabular}
    \end{center}
    \caption{Average running time of different algorithms in seconds on test datasets.}
    \label{fig:run_time}
\end{table}

\subsection{Running time:}
We train the GNN for each of the sparsification problems. For the Steiner tree and subsetwise multiplicative spanner problem, we train on geometric instances having 100 nodes. The training times are 20.48 and 21.29 hours respectively. For the subsetwise additive spanner problem, we train on geometric instances having 50 nodes. The training time is 4.92 hours. 
The average running times of the optimal algorithm, existing approximation algorithms, and our algorithm for different test datasets are shown in Table~\ref{fig:run_time}. We denote the Steiner tree, multiplicative spanner, and additive spanner problem instances by ST, MS, and AS respectively. These acronyms are followed by the number of nodes. All of these instances are geometric except the ST 80 dataset which represents the SteinLib 1080 dataset. We can see in Table~\ref{fig:run_time} that the approximation algorithms (Approx.)  are the fastest algorithms. Our algorithm is a little slower, however, the solution values are closer to the optimal values. 

\subsection{Impact of GNN prediction:}\label{sec:impact_GNN}
\begin{figure}[ht]
    \centering
    \includegraphics[width=.6\linewidth]{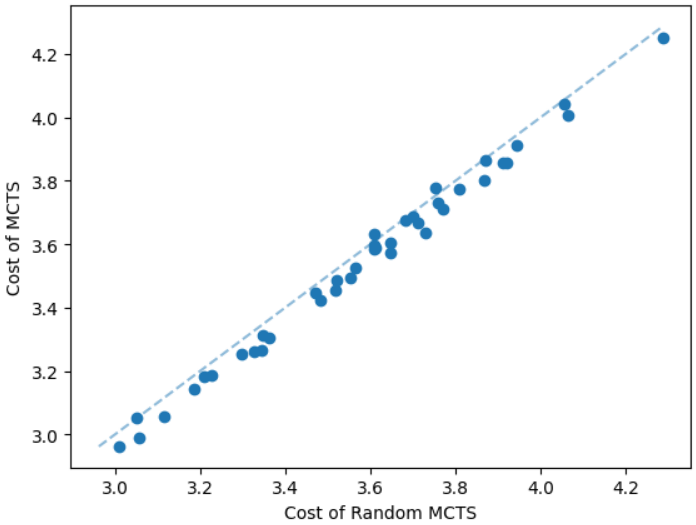}
    \caption{A comparison of our MCTS with another MCTS that selects random nodes from $\overline{S}$ (Random MCTS). This comparison illustrates the importance of GNN prediction. This is a dataset of geometric Steiner tree instances having 50 nodes.}
    \label{fig:gnn_mcts_random_mcts}
\end{figure}

The performance of the MCTS depends on the prediction of GNN. The task of GNN is to predict an important node $u$ from $\overline{S}$ in each step. This non-terminal node $u$ should connect the terminal nodes in such a way that overall the cost of the sparsification gets reduced. We provide a simple comparison to indicate the importance of the GNN prediction. We compare the MCTS that uses the GNN prediction with another MCTS method that selects random non-terminal nodes (Random MCTS) without using the GNN prediction. We take a dataset of geometric Steiner tree instances having 50 nodes. We illustrate the comparison in Figure~\ref{fig:gnn_mcts_random_mcts}. Our MCTS method computes Steiner trees with lower costs as expected.

\subsection{Performance on larger instances:}

\begin{figure}[ht]
    \centering
    \includegraphics[width=.6\linewidth]{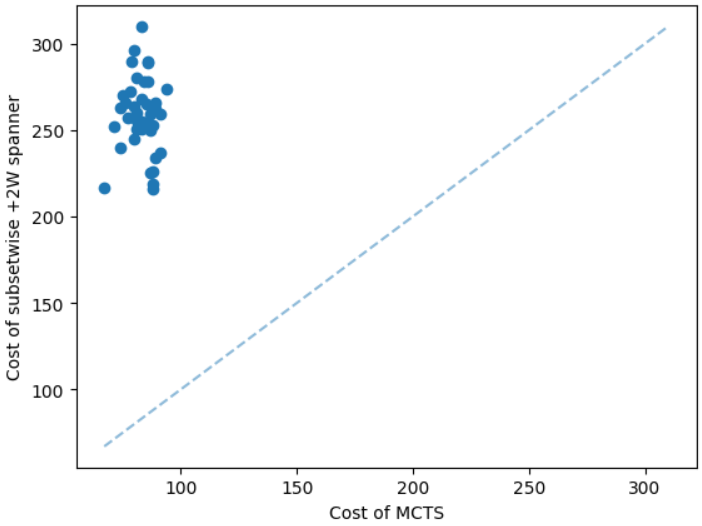}
    \caption{Performance on random geometric additive spanner instances ($\beta = 2W$). Each of these instances has a hundred nodes. Our algorithm (MCTS) is significantly better than the subsetwise +2W spanner.}
    \label{fig:GE100_GNN_APX_additive}
\end{figure}

We test our model on instances larger than the training instances to show the scalability of the model. 
We provide some results in the appendix due to the page limit.
For the additive spanner problem, we train the GNN on instances having 50 nodes. 
We are unable to compute an optimal solution for instances having 100 nodes due to the time limit. 
However, the MCTS can find a solution only in a few minutes. The solution computed by the MCTS is significantly better than the subsetwise +2W spanner algorithm, see Figure~\ref{fig:GE100_GNN_APX_additive}. The average running times of subsetwise +2W spanner algorithm and the MCTS are 14.82 and 79.13 seconds respectively.

\section{Conclusion}
We have described an approach for different sparsification problems based on GNNs and MCTS. An experimental evaluation shows that the proposed method computes solutions that are closer to optimal solutions on different datasets in a reasonable time. The proposed method is a generalization of the approximation algorithms and never performs worse than the approximation algorithms. 
The source code and experimental data can be found on \href{https://github.com/abureyanahmed/gnn-msts-sparsification}{GitHub}. 

\bibliographystyle{plain}
\bibliography{references}

\newpage

\appendix
\addcontentsline{toc}{section}{Appendices}
\section*{Appendix}

\section{GNN architecture:}

\begin{figure}[ht]
    \centering
    \includegraphics[width=.6\linewidth]{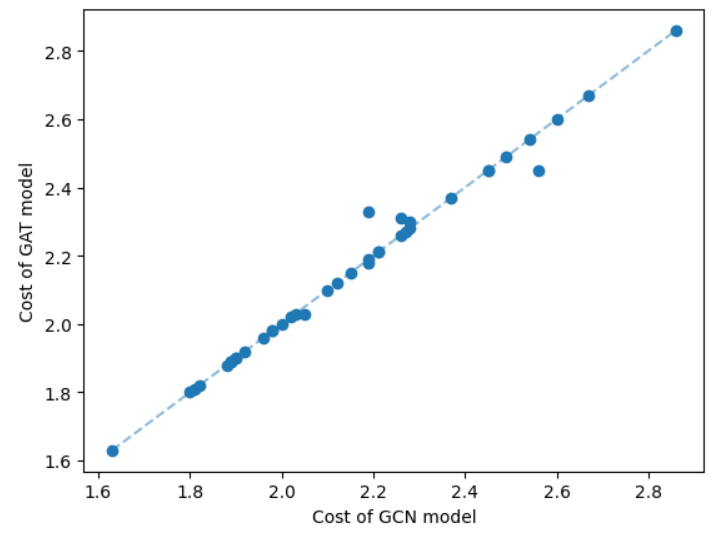}
    \caption{A comparison of GCN and GAT message-passing framework on geometric Steiner instances having 50 nodes.}
    \label{fig:gcn_gat}
\end{figure}

One key component of the GNN model is the message-passing procedure that aggregates information from the local neighborhood. We have incorporated two common message-passing procedures: graph convolutional network (GCN)~\cite{kipf2016semi} and graph attention network (GAT)~\cite{velivckovic2017graph}. We show a comparison of these two networks on a set of geometric Steiner tree instances having 50 nodes in Figure~\ref{fig:gcn_gat}.

Since the cost difference between the two types of message-passing models is not significant, we set the GCN model as the default.

\section{Experimental results}

\begin{figure*}[ht]
\minipage{0.32\textwidth}
  \includegraphics[width=\linewidth]{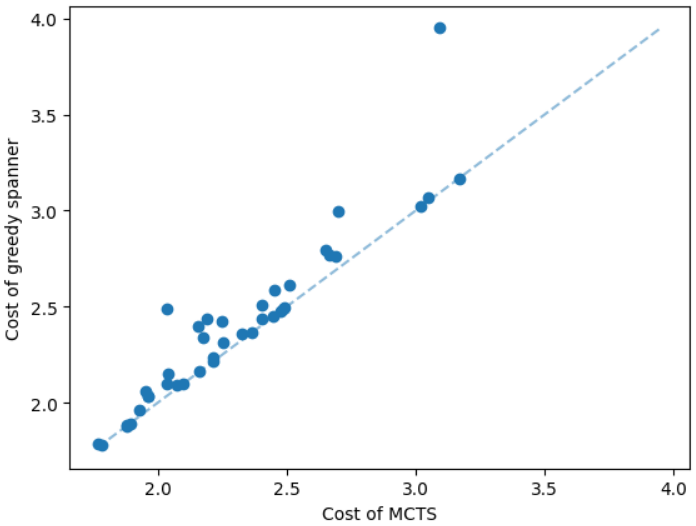}
  \subcaption{Twenty nodes graphs}
  \label{fig:GE20_GNN_APX_spanner}
\endminipage\hfill
\minipage{0.32\textwidth}
  \includegraphics[width=\linewidth]{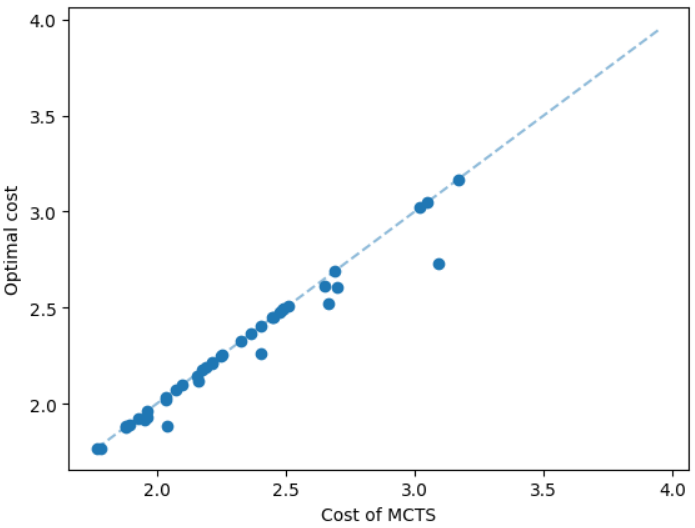}
  \subcaption{Twenty nodes graphs}
  \label{fig:GE20_GNN_OPT_spanner}
\endminipage\hfill
\minipage{0.32\textwidth}
  \includegraphics[width=\linewidth]{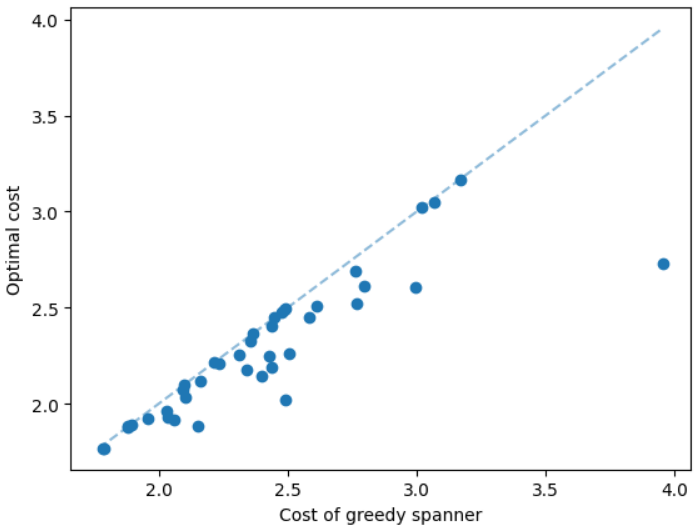}
  \subcaption{Twenty nodes graphs}
  \label{fig:GE20_APX_OPT_spanner}
\endminipage\hfill

\minipage{0.32\textwidth}
  \includegraphics[width=\linewidth]{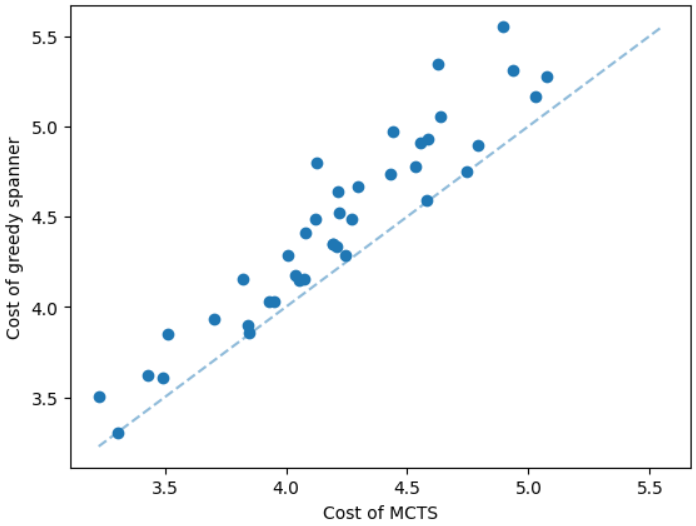}
  \subcaption{Fifty nodes graphs}
  \label{fig:GE50_GNN_APX_spanner}
\endminipage\hfill
\minipage{0.32\textwidth}
  \includegraphics[width=\linewidth]{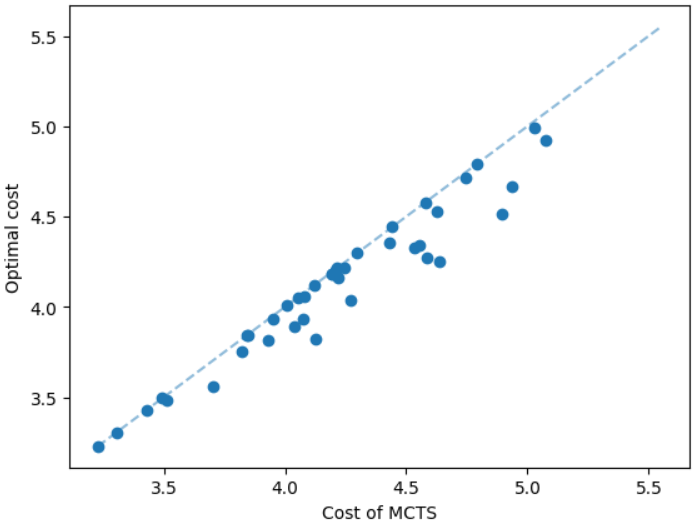}
  \subcaption{Fifty nodes graphs}
  \label{fig:GE50_GNN_OPT_spanner}
\endminipage\hfill
\minipage{0.32\textwidth}
  \includegraphics[width=\linewidth]{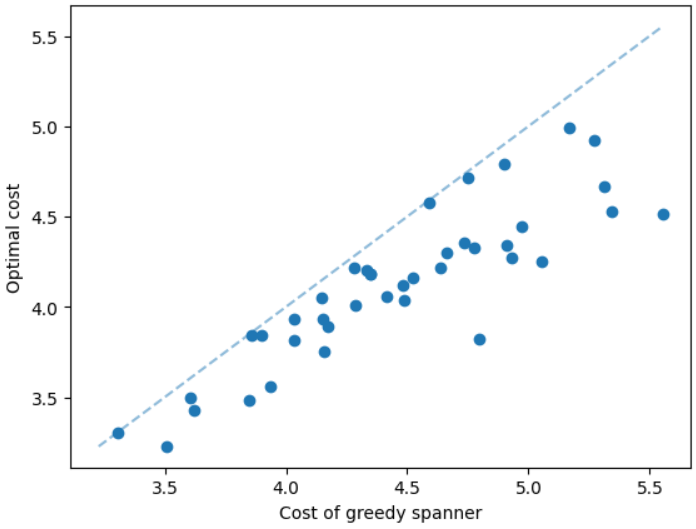}
  \subcaption{Fifty nodes graphs}
  \label{fig:GE50_APX_OPT_spanner}
\endminipage\hfill

\minipage{0.32\textwidth}
  \includegraphics[width=\linewidth]{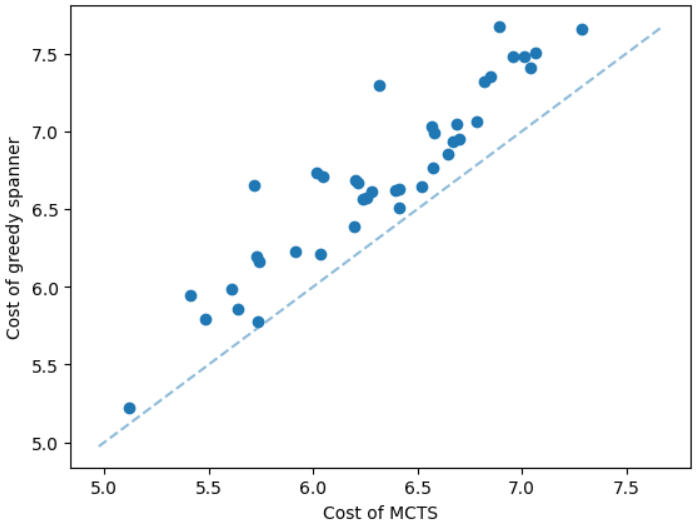}
  \subcaption{Hundred nodes graphs}
  \label{fig:GE100_GNN_APX_spanner}
\endminipage\hfill
\minipage{0.32\textwidth}
  \includegraphics[width=\linewidth]{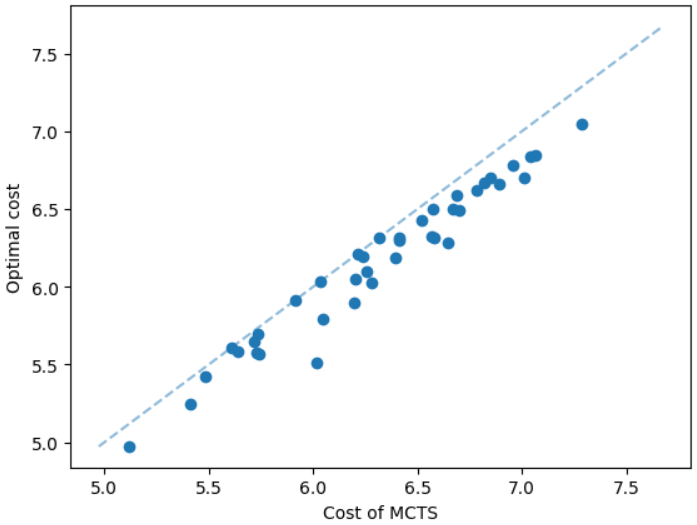}
  \subcaption{Hundred nodes graphs}
  \label{fig:GE100_GNN_OPT_spanner}
\endminipage\hfill
\minipage{0.32\textwidth}
  \includegraphics[width=\linewidth]{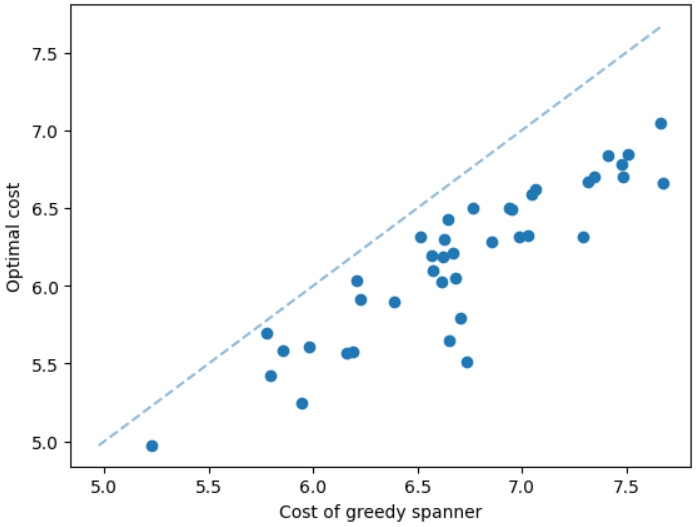}
  \subcaption{Hundred nodes graphs}
  \label{fig:GE100_APX_OPT_spanner}
\endminipage\hfill

\caption{Performance on random geometric multiplicative spanner instances ($\alpha = 2$). The lower the cost the better the algorithm is. Our algorithm (MCTS) performs better than the greedy algorithm.}
\label{fig:performance_GE_spanner}
\end{figure*}

We now consider the subsetwise multiplicative spanner problem. Here, the multiplicative stretch $\alpha$ is equal to 2. Similar to the Steiner tree problem, we train the GNN on geometric instances having 100 nodes and test the MCTS on instances having 20, 50, and 100 nodes. Here, we compare our method with the well-known greedy algorithm and an optimal algorithm. The greedy algorithm produces asymptotically tight spanners assuming the Erd\H{o}s girth conjecture and performs well in practice~\cite{spannersurvey}. The results are illustrated in Figure~\ref{fig:performance_GE_spanner}. We can see that the MCTS performs significantly well compared to the greedy algorithm for instances having different numbers of nodes. Also, the cost of MCTS is comparable with the optimal cost.

\section{Impact of sample size and height of the search tree:}
\begin{figure}[ht]
    \centering
    \includegraphics[width=.6\linewidth]{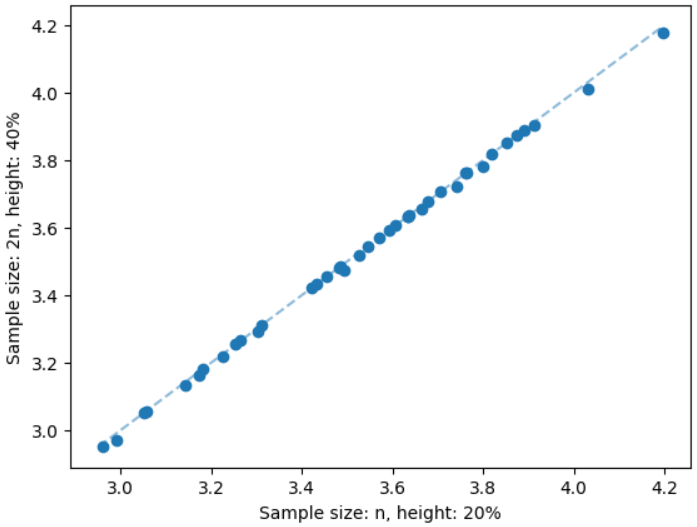}
    \caption{A comparison of the MCTS using a sample size equal to $n$ and the height of the search tree equal to 20\% of $n$ with a sample size equal to $2n$ and the height of the search tree equal to 40\% of $n$. Here, $n$ is the number of nodes in the input graph. The MCTS gets saturated after using the first setting and does not provide a significant improvement with the second setting. This is a dataset of geometric Steiner tree instances having 50 nodes.}
    \label{fig:GE50_samples}
\end{figure}

The sample size and height of the search tree are important parameters of the MCTS. We keep the sample size equal to $n$, where $n$ is the number of nodes of the input graph. 
We stop the MCTS when the height of the search tree is equal to 20\% of $n$.
The reasons for keeping the height only 20\% of $n$ are the computational cost per iteration and the effectiveness of the GNN prediction as discussed in Section~\ref{sec:impact_GNN}. For each sample node, we need to compute a single source shortest path that increases the total computational cost of the MCTS. Also, the GNN predicts most of the important nodes by the initial set of samples and after that, the sampling process gets saturated and does not increase the solution quality that much. For example, we illustrate the impact of a larger sample size and height of the search tree on geometric instances of the Steiner tree problem having 50 nodes in Figure~\ref{fig:GE50_samples}. We can see that the solution quality does not improve that much when we increase the sample size from $n$ to $2n$ and the height of the search tree from 20\% to 40\%. However, we significantly increase the solution quality after keeping the sample size equal to $n$ and the height of the search tree equal to 20\% of $n$, see Figure~\ref{fig:GE50_APX_OPT}. On the other hand, the average running time of the MCTS with the first setting is 3.90 seconds. The average running time increases to 7.13 seconds when we increase the sample size and height of the search tree. For each of the settings studied in this paper, we have found that a sample size equal to $n$ and the height of the search tree equal to 20\% of $n$ is enough. Hence we use this setting  for all experiments.

\section{Performance on larger instances:}

\begin{figure*}[ht]

\minipage{0.32\textwidth}
  \includegraphics[width=\linewidth]{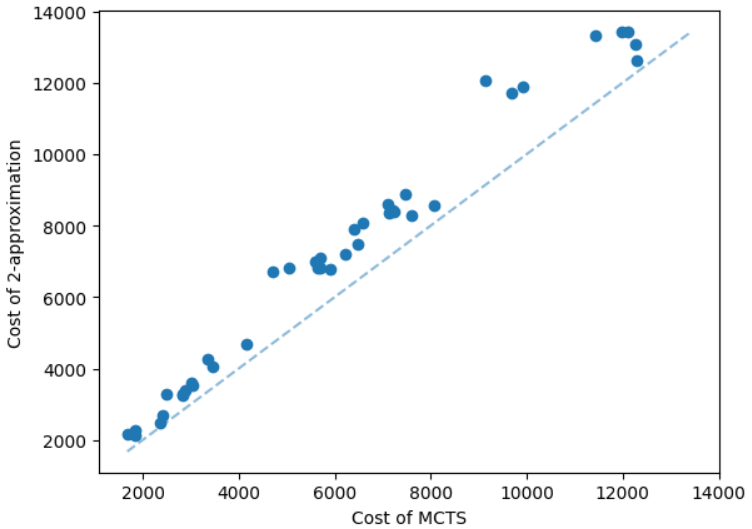}
  \label{fig:I160_GNN_APX}
\endminipage\hfill
\minipage{0.32\textwidth}
  \includegraphics[width=\linewidth]{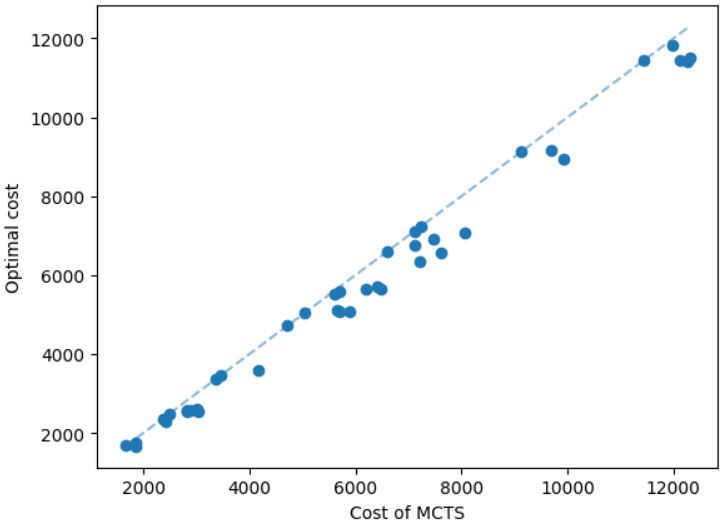}
  \label{fig:I160_GNN_OPT}
\endminipage\hfill
\minipage{0.32\textwidth}
  \includegraphics[width=\linewidth]{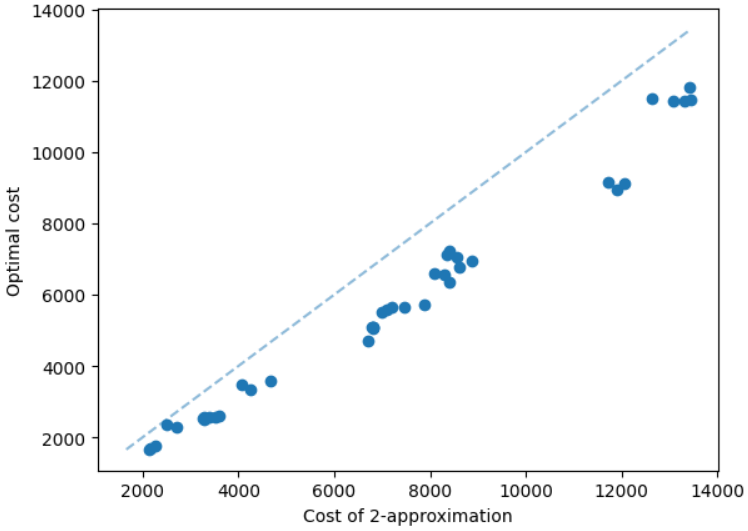}
  \label{fig:I160_APX_OPT}
\endminipage\hfill

\caption{Performance on SteinLib I160 dataset. The lower the cost the better the algorithm is. Our algorithm (MCTS) performs better than 2-approximation.}
\label{fig:I160}
\end{figure*}

For the Steiner tree problem, we train the GNN on instances having 100 nodes. In an earlier section, we compared our method with instances having 100 or fewer nodes. We now compare our method to larger instances. In Figure~\ref{fig:I160}, we illustrate the performance of our method on SteinLib I160 graphs; each of these instances contains 160 nodes. These results indicate that our method performs well on larger instances even after training on small instances.

\end{document}